\documentclass{article} 
\usepackage{iclr2026_conference,times}

\usepackage{amsmath,amsfonts,bm}

\def\eqref#1{equation~\ref{#1}}

\def\1{\bm{1}}

\DeclareMathAlphabet{\mathsfit}{\encodingdefault}{\sfdefault}{m}{sl}
\SetMathAlphabet{\mathsfit}{bold}{\encodingdefault}{\sfdefault}{bx}{n}

\usepackage{hyperref}
\usepackage{url}
\usepackage{wrapfig}
\usepackage{float}
\usepackage{caption} 
\usepackage{graphicx}
\usepackage{listings}
\usepackage{booktabs}
\usepackage{subcaption}
\usepackage{pifont}
\usepackage{array}
\usepackage{tabularx}
\usepackage{longtable}
\usepackage{fontawesome5}
\usepackage[most]{tcolorbox}
\usepackage{enumitem}
\usepackage{multirow}

\definecolor{InteractionGreen}{HTML}{299B8D}
\definecolor{InteractionBlue}{HTML}{417CB9}
\definecolor{InteractionAmber}{HTML}{DD8A3C}
\colorlet{InteractionGreenBg}{InteractionGreen!8!white}
\colorlet{InteractionBlueBg}{InteractionBlue!8!white}
\colorlet{InteractionAmberBg}{InteractionAmber!8!white}
\newtcolorbox{interactionprompt}[2][InteractionGreen]{%
  enhanced, breakable, lines before break=5,
  colback=#1Bg,
  colframe=#1,
  colbacktitle=#1,
  coltitle=white,
  title={#2},
  fonttitle=\bfseries,
  fontupper=\small,
  boxrule=1.0pt, arc=3pt,
  left=8pt, right=8pt, top=8pt, bottom=8pt,
  attach boxed title to top center={yshift=-2mm},
  before skip=6pt, after skip=6pt,
  before upper={\setlength{\parindent}{0pt}\setlength{\parskip}{2pt}},
  overlay middle and last app={\node[above,inner sep=0pt,outer sep=0pt] at ([yshift=-2mm]frame.north)
    {\tcbox[enhanced,colframe=#1,colback=#1,boxrule=0.5mm,arc=1mm,boxsep=1mm,left=2mm,right=2mm,top=0mm,bottom=0mm]{\color{white}\bfseries #2\space (continued)}};}
}

\ifcsname c@rtbenchprompt\endcsname\else
  \newcounter{rtbenchprompt}
\fi

\title{Realtime-Bench: Can Frontier Models Solve Real-time Interactive Problems?}

\author{%
\textbf{Jingyu Tang$^{1,*}$, Xu Yang$^{1,*}$, Zhehao Zhang$^{2,*}$, Zheng Wang$^{3}$, Qi Cheng$^{4}$, Haozhe Zhao$^{3}$}, \\
\textbf{Manling Li$^{5}$, Wenxuan Li$^{6}$, Xide Xia$^{6}$, Jiaxin Pei$^{1,\dagger}$} \\[0.5em]
\normalfont\small $^{1}$University of Texas at Austin,
\normalfont\small $^{2}$The Ohio State University,
\normalfont\small $^{3}$University of Illinois at Urbana-Champaign \\
\normalfont\small $^{4}$NVIDIA \quad $^{5}$Northwestern University \quad $^{6}$Steadyworks \\[0.3em]
\normalfont\small $^{*}$Equal contribution. \quad $^{\dagger}$Corresponding author. \\
\normalfont\small \texttt{jingyu@utexas.edu},
\texttt{jiaxinpei@utexas.edu} \\[0.5em]
\normalfont\small
\href{https://realtimeagents.org/\#/home}{\faGlobe\enspace Project Website}\qquad
\href{https://github.com/RealtimeAgents/Realtime-Bench}{\faGithub\enspace Code}
}

\usepackage{xcolor}
\definecolor{linkblue}{RGB}{0,90,200}

\newcommand{\benchmark}{Realtime-Bench}

\iclrfinalcopy 
\begin{document}

\maketitle

\vspace{-0.8cm}
\begin{abstract}

Real-time interaction requires agents to perceive evolving environments and act under temporal constraints, yet existing evaluations of real-time streaming models focus on conversational quality and streaming perception, leaving effective action under strict temporal constraints untested. We introduce Realtime-Bench, a suite of dynamic tasks where delayed actions directly affect success. Under Wall Clock, where environments advance according to actual inference latency, the best success rate reaches only 44\% and the best Pass\textasciicircum{}3 is only 36\%, demonstrating that real-time interaction remains challenging. To separate serving latency from models’ real-time streaming capability, we compare Wall Clock with Paused Clock, which removes inference delays as an idealized reference, and introduce Token Clock, which advances simulated time according to generated token counts, testing temporal awareness independently of serving speed. We find that temporal costs limit success in evolving environments and that higher reasoning effort has mixed effects under time pressure, reflecting a trade-off between decision quality and timely execution. Based on these results, we discuss future directions for improving real-time multimodal models, including adaptive reasoning with temporal awareness and continuous perception and action. To facilitate future research, we release the dataset and host a public leaderboard at \href{https://realtimeagents.org}{\textcolor{linkblue}{\texttt{realtimeagents.org}}}.

\end{abstract}

\begin{figure}[H]
    \centering
    \vspace{-8pt}
    \includegraphics[width=\linewidth]{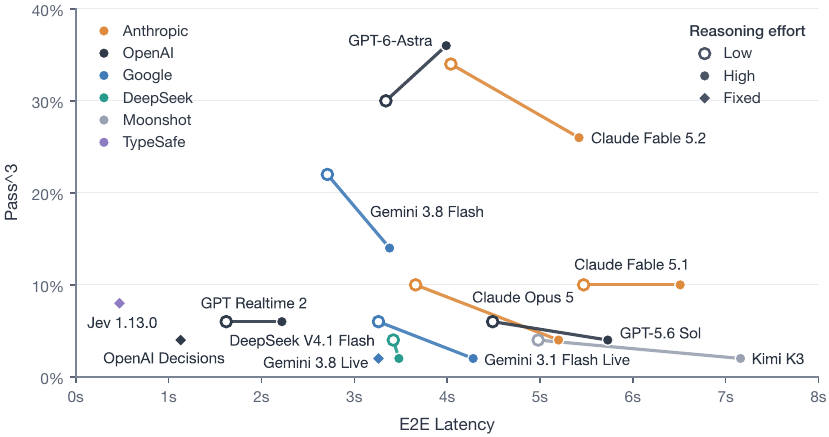}
    \caption{\textbf{Performance of Frontier Models on Realtime-Bench.} Pass\textasciicircum{}3 measures the proportion of tasks successfully completed in all three independent trials, capturing the reliability needed in time-sensitive scenarios where agents often have only one chance to succeed.}
    \label{fig:latency-pass3}
\end{figure}

\section{Introduction}

Many real-world interactive scenarios require decisions to be made under strict temporal constraints. For example, when purchasing a highly demanded ticket, detecting availability is not enough: an agent must observe the changing interface and act before the opportunity disappears. Such scenarios require \emph{real-time streaming capabilities}, where a model continuously perceives an evolving environment and produces timely actions in response to newly arriving information. Recently, a growing family of real-time streaming models, such as GPT-Realtime~\citep{openai2025gptrealtime} and Gemini Live~\citep{IntroducingGemini382026}, has begun to make this setting practically possible. In contrast to asynchronous models that process inputs only after they are fully received, these models can consume streaming inputs and react to new information with low latency. 

Unlike previous benchmarks that evaluate models in static environments~\citep{zhang2026videogamebenchvisionlanguagemodelscomplete, tan2026stardojobenchmarkingopenendedbehaviors} where the environment does not continuously evolve, evaluating streaming models requires dynamic settings that test whether models can perceive and act effectively under strict temporal constraints. 
However, existing evaluations of real-time systems largely focus on capabilities such as streaming perception, conversational quality, and interruption handling~\citep{11463959, lu2026omniinteractbenchmarkingrealworldstreaming}, with limited coverage of interactive tasks in which actions affect subsequent environment states and delayed execution can change the outcome.

We first characterize \emph{real-time streaming capability} along three complementary dimensions: temporal state awareness, adaptive deliberation, and time-aware action selection. These dimensions capture an agent's ability to track relevant changes and information freshness, adjust deliberation to the available time, and select actions that remain useful when executed. Building on this framework, we introduce \textbf{\benchmark{}}, comprising 50 interactive game and web tasks in which evolving states and limited opportunities make success depend on both action choice and timing.

We evaluate 12 frontier models spanning standard asynchronous models, streaming models and decision models. As shown in Figure~\ref{fig:latency-pass3}, real-time interaction remains challenging, with the best success rate reaching only 44.0\% and best pass\textasciicircum{}3 is 36\% under \textbf{Wall Clock}, which advances the environment according to actual inference and serving time, following prior interactive-agent evaluations~\citep{NEURIPS2025_ddaec864, froger2026gaia, ValsAI, parekh2026gptntbenchmarkingrealtimecollaboration}. 

\begin{figure*}[hp]
\centering
\includegraphics[width=0.98\linewidth]{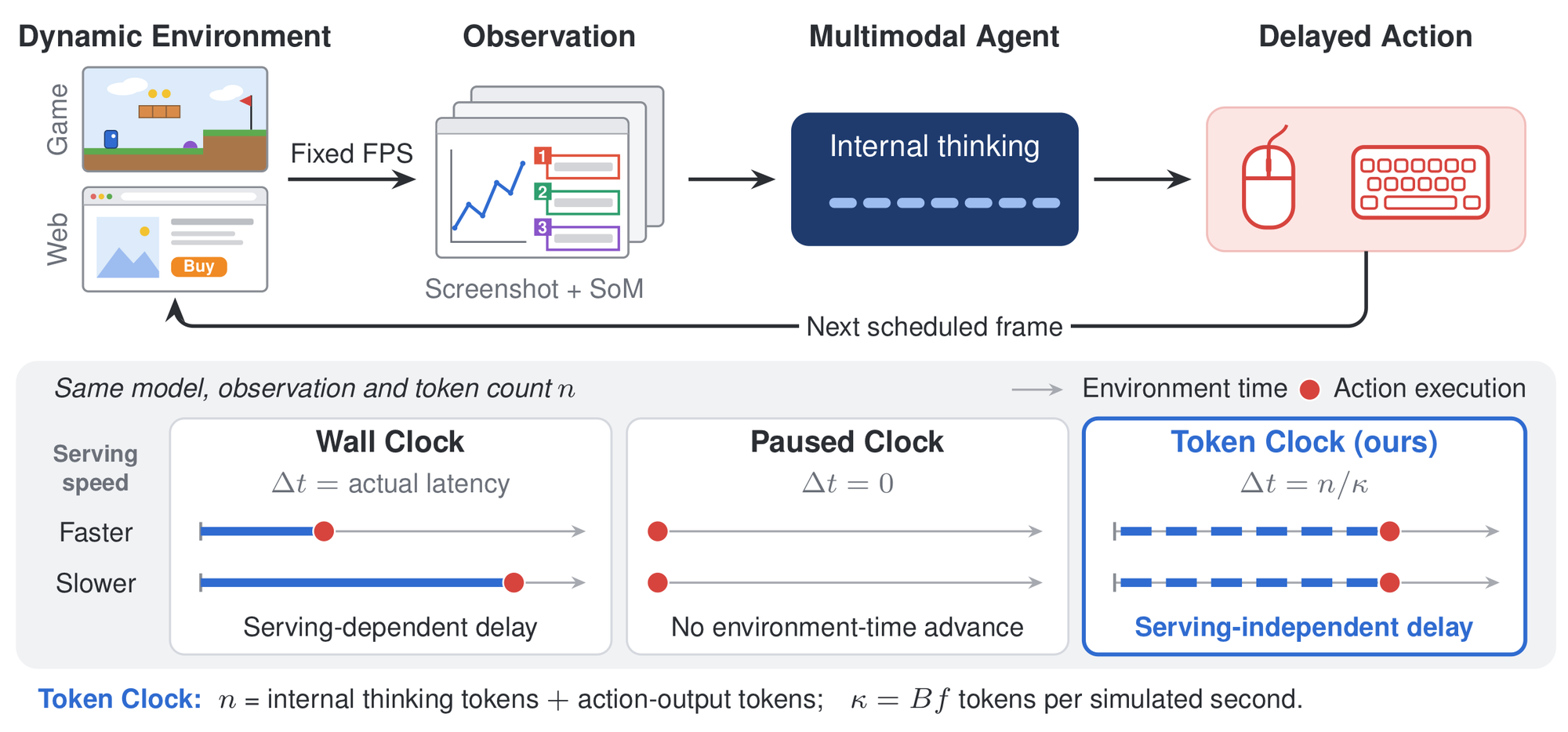}
\caption{\textbf{Realtime-Bench evaluates models on interactive game and web tasks under three clock protocols.} Wall Clock advances environments with actual inference latency, while Paused Clock removes inference delays as an idealized reference. \textbf{Token Clock (ours)} advances simulated time according to generation tokens consumed before action commitment, testing temporal awareness of models independently of physical serving speed.}
\label{fig:setting}
\end{figure*}
To better understand these limitations and disentangle the effects of serving latency and models' real-time streaming capability, we compare three clock protocols (Figure~\ref{fig:setting}). Alongside Wall Clock, \textbf{Paused Clock} freezes the environment during inference, providing an idealized reference in which deliberation incurs no temporal cost~\citep{tan2026stardojobenchmarkingopenendedbehaviors, zheng-etal-2026-v, 3780338.3782666}. We further introduce \textbf{Token Clock}, which advances simulated time according to the number of generation tokens consumed before an action is committed. Models that spend more tokens deliberating therefore act later, preserving the task consequences of delayed decisions while controlling for physical serving speed. Under each clock, we test low and high reasoning effort to examine the trade-off between decision quality and timely execution. Our analysis reveals that substantial performance gaps relative to Paused Clock persist under Token Clock, highlighting the difficulty of translating task competence into timely action even when temporal costs are standardized independently of serving speed. Higher reasoning effort improves success for all six models under Paused Clock but has mixed effects under temporal pressure. Further experiments show that relaxing the temporal cost of generation generally improves performance but does not ensure task success. These findings motivate future work on adaptive reasoning with temporal awareness and continuous perception and action, enabling agents to coordinate deliberation and execution with the time available.

\section{Related Work}

\subsection{Streaming Multimodal Models and Benchmarks}

Video multimodal models have progressively evolved from offline video understanding~\citep{maaz-etal-2024-video,lin-etal-2024-video,liu2024llavanext,Liu_2024_CVPR,liu2023llava} toward real-time streaming interaction, where models continuously perceive incoming observations and decide when and how to respond as the stream unfolds~\citep{Chen_2024_CVPR,wang-etal-2025-videollm,11094534,cui2026minicpmo45realtimefullduplex}. 
Existing training and evaluation formulate real-time interaction as responding to a continuously arriving but sparsely sampled audiovisual stream, commonly at around 1--2 FPS~\citep{Chen_2024_CVPR,11094534,proactvl,lu2026omniinteractbenchmarkingrealworldstreaming}. Under this setting, existing benchmarks focus on evaluating the abilities of streaming perception~\citep{11463959,11093281,river}, response timing~\citep{wang2025proactivevideoqacomprehensivebenchmarkevaluating,he2026omniduplexevalevaluatingrealtimeduplex,lu2026omniinteractbenchmarkingrealworldstreaming}, interruption handling~\citep{11433838,lu2026omniinteractbenchmarkingrealworldstreaming}, and conversational quality~\citep{he2026omniduplexevalevaluatingrealtimeduplex,lu2026omniinteractbenchmarkingrealworldstreaming} in scenarios such as surveillance anomaly detection~\citep{wang2025proactivevideoqacomprehensivebenchmarkevaluating}, question answering during gym exercises and shopping, and step-by-step task guidance~\citep{lu2026omniinteractbenchmarkingrealworldstreaming}, leaving untested whether models can act effectively in closed-loop environments where the timing of their decisions affects subsequent states and task outcomes. In contrast, \benchmark{} evaluates closed-loop interaction in web environments and games, where task success depends on both what actions models take and when they take them.

\subsection{Real-Time Evaluation in Interactive Environments}

Existing interactive evaluations assess asynchronous vision-language agents in games that pause during model inference, allowing their perception, reasoning, and planning abilities to be evaluated independently of response speed~\citep{3780338.3782666,zhang2026videogamebenchvisionlanguagemodelscomplete,tan2026stardojobenchmarkingopenendedbehaviors,zheng-etal-2026-v,ouyang2026gameworldstandardizedverifiableevaluation}. To evaluate agents under real-time pressure, VideoGameBench~\citep{zhang2026videogamebenchvisionlanguagemodelscomplete}, GameWorld-RT~\citep{ouyang2026gameworldstandardizedverifiableevaluation},  GPTNT~\citep{parekh2026gptntbenchmarkingrealtimecollaboration}, and Gaia2~\citep{froger2026gaia} allow game or web environments to continue evolving during inference, making task success depend on both decision quality and wall-clock latency. Although such wall-clock capture the consequences of slow responses, they also tie evaluation to deployment-specific inference speed. Similar to Real-Time Reasoning Gym~\citep{ICLR2026_ccbe1604}, we adopt Token Clock, which maps generated tokens to environment time at a shared rate, preserving the temporal cost of deliberation while decoupling evaluation from physical serving speed.

\section{Setting Formulation}

\subsection{Real-Time Streaming Capabilities}

We define real-time streaming capability as the ability to perceive, reason, and act effectively while the environment continues to evolve. We characterize this capability along three complementary dimensions. (1) \textit{Temporal state awareness (RT-Perceive)} requires models to use available observations to track relevant changes, information freshness, and opportunity timing; (2) \textit{Adaptive deliberation (RT-Deliberate)} requires models to adjust their reasoning length and depth according to the available time and the expected benefit of further computation; (3)\textit{Time-aware action selection (RT-Act)} requires models to determine which action will remain useful when executed, including whether to wait or act immediately and what action to take.

These dimensions jointly shape how an agent responds to an evolving environment. For example, an agent purchasing a ticket must track the sale status and assess whether observed availability is still current \textit{(RT-Perceive)}, limit further deliberation when the purchase window is closing \textit{(RT-Deliberate)}, and choose between waiting for sales to open and purchasing while tickets remain available \textit{(RT-Act)}.

\subsection{Clock Protocols}




As shown in Figure~\ref{fig:setting}, we use three clock protocols. \textbf{Wall Clock} advances the environment by the actual inference time, so action timing depends on both model computation and serving infrastructure. \textbf{Paused Clock} freezes the environment during generation, providing an idealized reference in which inference incurs no temporal cost. \textbf{Token Clock} advances the environment at a fixed rate per generation token, so consuming more tokens delays execution regardless of actual serving latency.

To formalize Token Clock, let $B$ denote the number of generation tokens corresponding to one observation interval and $f$ the observation rate in frames per simulated second. The token-to-time rate is $\kappa = Bf$, measured in generation tokens per simulated second. If the model generates $n$ tokens after receiving an observation and before committing to an action, the action is applied after a simulated-time delay of
\[
    \Delta t = \frac{n}{\kappa} = \frac{n}{Bf}.
\]
In both Token Clock and Wall Clock, when generation spans multiple observation intervals, intermediate frames are skipped, and the model receives the most recent scheduled frame available at the resulting simulated time. This provides a simplified analogue of humans returning attention to the current scene after deliberation, and tests whether the model can respond before relevant events or action opportunities pass. Both protocols thus require the model to balance the benefits of additional deliberation against the cost of delayed action.

\begin{wrapfigure}{r}{0.42\textwidth}
\centering
\vspace{-80pt}
\includegraphics[width=1\linewidth]{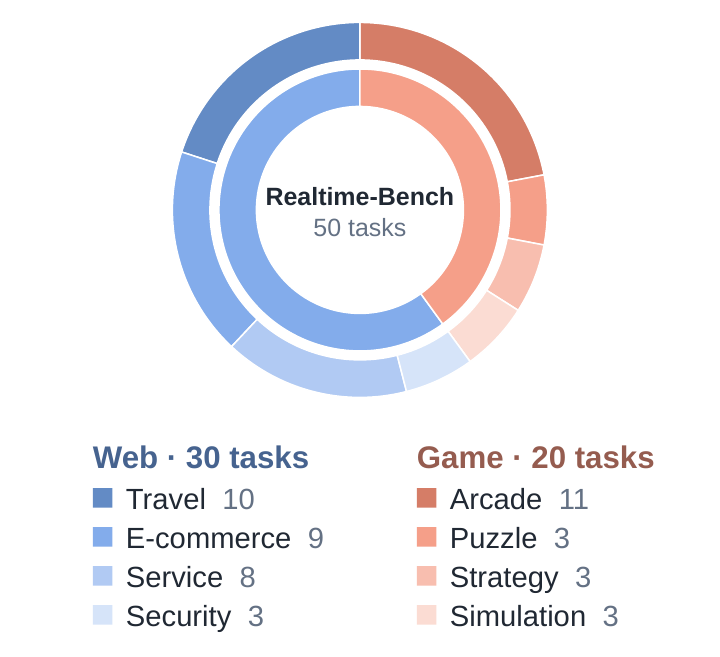}
\caption{Task distribution in Realtime-Bench. 
}
\vspace{-20pt}
\label{fig:app-task-distribution}
\end{wrapfigure}

\section{Benchmark Design}

\subsection{Design principles.}
To evaluate models' real-time streaming capabilities in environments that continue to evolve while they observe, reason, and act, we design tasks around the following three criteria.

\textbf{Real-world fidelity.} To capture practical real-time interaction demands, task interfaces and interaction mechanics should reflect realistic applications and games. For example, web tasks simulate real-world websites such as Amazon, while game tasks feature popular games.

\textbf{Dynamically evolving environment.} Environment states should change over time, independently of the model's actions. These changes may include moving game elements, changing availability, or continuously updated information. For example, available seats change during cinema booking, while obstacles move across the screen in games.

\textbf{Temporally constrained interaction.} Success should depend on both what actions the model takes and when it takes them. Deadlines, limited action windows, and transient opportunities require timely responses. For example, a model must reserve a seat before it becomes unavailable or dodge an obstacle before a collision.

\subsection{Task construction.}
Guided by the above criteria, we construct web tasks based on realistic application scenarios and adapt game tasks from popular open-source games. We adjust task goals and temporal constraints so that success requires timely action, while ensuring that tasks remain feasible for humans. This process yields 50 tasks: 30 web tasks and 20 game tasks. Each task has task-specific success criteria, such as successfully purchasing a ticket before the countdown expires.

\subsection{Task examples and distribution.}
\begin{figure*}[h]
    \centering
\includegraphics[width=0.98\linewidth]{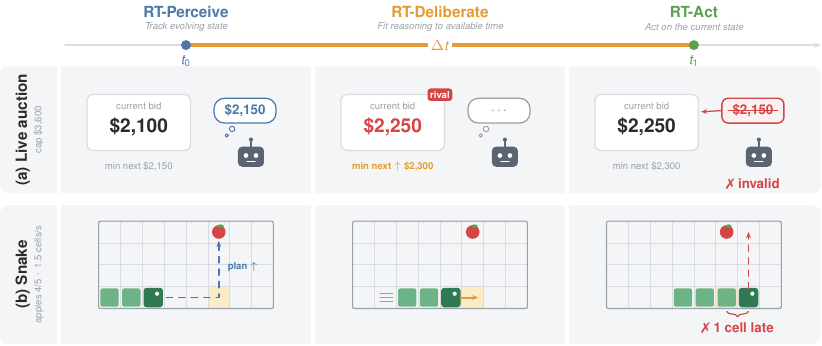}

    \caption{Representative tasks in Realtime-Bench. (a) Live auction (web): win a live auction under a \$3,600 cap. (b) Snake (game): steer a moving snake to eat five apples. Every task in our benchmark tests three real-time streaming capabilities: RT-Perceive, RT-Deliberate, and RT-Act.}
    \label{fig:task_examples}
\end{figure*}

Figure~\ref{fig:task_examples} illustrates two representative tasks, one from each domain: the web task requires placing a winning bid in a live auction before the bidding window closes, while the game task requires steering a moving snake to eat five apples. Figure~\ref{fig:app-task-distribution} shows the distribution of task scenarios.

\section{Evaluation Protocol}
\label{sec:evaluation_protocol}

\subsection{Observation and Action Interface}
\label{sec:observation_action_interface}

The environment provides observations at a fixed rate of $f$ frames per simulated second.
At each scheduled observation frame, the model receives the current screenshot
together with the retained interaction history. To prevent GUI grounding from
becoming a bottleneck for evaluating real-time decision making, we augment the
screenshot with a read-only Set-of-Marks (SoM) representation following common
GUI-agent practice~\citep{yang2023setofmarkpromptingunleashesextraordinary}, which identifies currently interactable elements and exposes their associated semantics and supported actions.
At each decision step, the model generates executable GUI actions from a common GUI action space (e.g., \texttt{click}, \texttt{type}, \texttt{scroll}, and \texttt{wait})~\citep{NEURIPS2024_5d413e48}. The detailed implementation
are provided in Appendix~\ref{app:interface}.

\subsection{Context Management}
\label{sec:history_alignment}

To minimize harness-induced overhead when evaluating models' real-time capabilities, we follow Qwen-CUA~\citep{lu2026qwencuanativecomputeruse} in removing older history in blocks rather than at every step. This keeps the history prefix unchanged between folds, allowing models to reuse its KV cache across consecutive steps and reducing repeated prefill computation caused by frequent history truncation. All models follow the same history retention policy. Implementation details are provided in Appendix~\ref{app:history}.

\subsection{Episode Execution}
\label{sec:episode_execution}

Suppose the model receives an observation at environment time $t_i$ and commits an action after a delay $\Delta t_i$ determined by the clock protocol. Under \textbf{Wall Clock}, $\Delta t_i$ equals the actual inference latency; under \textbf{Paused Clock}, $\Delta t_i = 0$; under \textbf{Token Clock}, $\Delta t_i$ is determined by the generation tokens consumed before action commitment. The action timestamp is
\[
t_i^{\mathrm{act}} = t_i + \Delta t_i.
\]
The environment advances from $t_i$ to $t_i^{\mathrm{act}}$ before the action is applied. If generation terminates without a valid action, the harness substitutes \texttt{wait} at the timestamp determined by the same clock protocol. GUI execution latency does not advance environment time.

An episode terminates when the environment reaches a task-defined terminal condition (e.g., game over, successful completion of a web transaction, or expiration of a task-specific countdown) or the benchmark-defined time horizon (e.g., five minutes of environment time). The scorer evaluates the terminal environment state against task-specific success criteria, such as completing the first level of a game or securing a ticket.

\section{Experiments}

\subsection{Experimental Setup}
\label{sec:exp_setup}

\paragraph{Models.}
We evaluate 12 frontier models spanning (1) standard asynchronous models, including GPT-6-Astra~\citep{GPT6AstraNew}, GPT-5.6 Sol~\citep{PreviewingGPT56Sol2026}, Claude Fable 5.1~\citep{IntroducingClaudeFable}, Claude Opus 5~\citep{IntroducingClaudeOpus}, Gemini 3.8 Flash~\citep{IntroducingGemini382026a}, Deepseek V4.1 Flash~\citep{IntroducingDeepSeekV41FlashSmarter2026} and Kimi-K3~\citep{team2026kimi}; (2) real-time streaming models, which are optimized for low-latency interaction, including GPT-Realtime-2~\citep{openai2025gptrealtime}, Gemini 3.1 Flash live~\citep{Gemini31Flash2026} and Gemini 3.8 Live~\citep{IntroducingGemini382026b};  and (3) decision models, which directly return structured decisions rather than generate free-form text, including OpenAI Decisions~\citep{DecisionsOpenAIAPI} and Jev 1.13~\citep{IntroducingSystemOne}.

\paragraph{Evaluation settings.}

To evaluate model performance in real-world deployment settings, we first evaluate all 12 models under Wall Clock. We report the pass rate for each run, defined as the fraction of tasks successfully completed, along with pass\textasciicircum{}k and pass@k for $k \in \{1,2,3\}$. Following $\tau$-bench~\cite{ICLR2025_1b126cc3}, pass\textasciicircum{}k measures the probability of succeeding in all $k$ attempts at a task, whereas pass@k measures the probability of succeeding in at least one of $k$ attempts. 

To further distinguish task performance under actual inference latency, without inference delays, and under a standardized temporal cost of token generation, we further compare Wall Clock, Paused Clock, and Token Clock on six representative models: GPT-Realtime-2, Gemini 3.1 Flash live, GPT-5.6 Sol, Claude Opus 5, Gemini 3.8 Flash, and Kimi-K3. We exclude decision models from this analysis because they do not report output token counts, which are required for Token Clock evaluation. For each clock protocol, we test both low and high reasoning effort to examine the tradeoff between decision quality and timeliness. Crossing the three clock protocols with the two reasoning levels yields six settings: WL and WH for Wall Clock, PL and PH for Paused Clock, and TL and TH for Token Clock, where L and H denote low and high reasoning effort, respectively. 

\paragraph{Temporal configuration.}
We set the observation rate to $f=1$ FPS, following the frame-sampling recipes primarily used in current training and evaluation of real-time models~\citep{Xia_2026_CVPR, 11094534, 11444205, qwenteam2026qwen35omnitechnicalreport}. For the main experiments of Token Clock, we set $B=64$, corresponding to $\kappa=Bf=64$ charged tokens per simulated second, a rate broadly consistent with practical autoregressive decoding speed~\citep{yang-etal-2026-longspec, comanici2025gemini25pushingfrontier}, which also suffice to produce a complete action together with one or two sentences of reasoning within one observation interval based on our pilot experiment. This provides room for brief deliberation that can support successful task execution, while longer reasoning can incur a temporal cost.

\paragraph{Human participants and procedure.}
We conducted a human study on Prolific with 27 participants, with each task attempted by five different participants using a mouse and keyboard on the same continuously updating interfaces. Details are provided in Appendix~\ref{app:human_study}.

\subsection{Overall Performance}

\paragraph{Performance remains limited under Wall Clock.}

As shown in Figure~\ref{fig:latency-pass3}, GPT-6 Astra (high) achieves the highest Pass\textasciicircum{}3 under Wall Clock at only 36\%, indicating that even the strongest evaluated model struggles to complete tasks consistently. Most models incur latencies exceeding one second. Jev-1.13.0 achieves sub-second latency  but its Pass\textasciicircum{}3 remains low at 8\%. These results suggest that low latency alone is insufficient for reliable real-time performance: models must combine timely responses with strong task-solving capabilities. Improving real-time abilities therefore requires advances in both response speed and model real-time streaming capability. Detailed per-run evaluation results, along with the full Pass\textasciicircum{}k and Pass@k results, are provided in the Appendix~\ref{app:performance_wall_clock}.

\begin{figure}[!htbp]
    \centering
    \includegraphics[width=\linewidth]{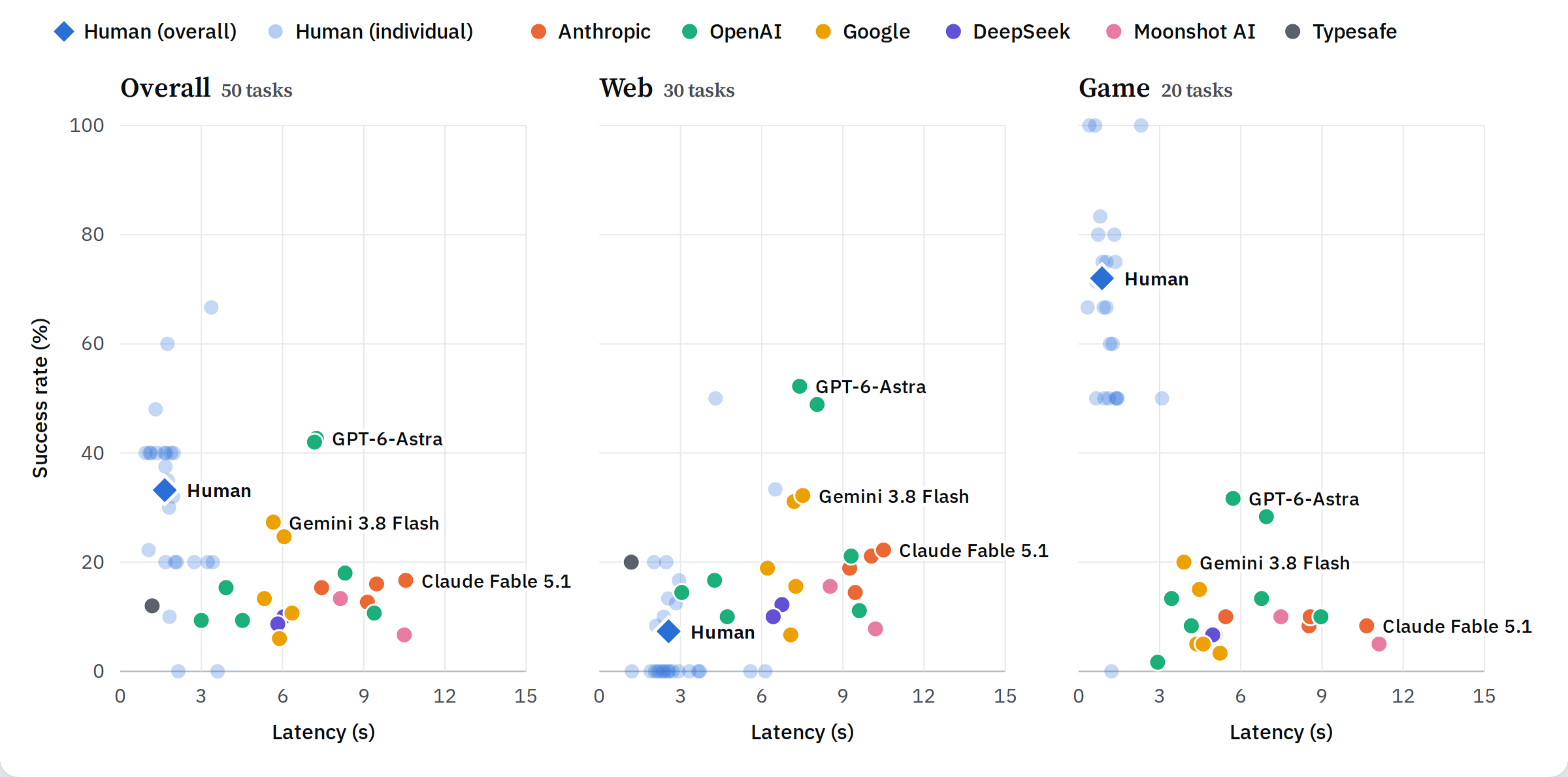}
    \caption{\textbf{Human and agent performance across task categories.} Success rate versus mean inter-action interval (labeled latency) overall, on web tasks, and on games. Diamonds denote aggregate human performance; light-blue points denote individuals.}
    \label{fig:human_evaluation}
\end{figure}

\paragraph{Humans and agents exhibit complementary strengths.}
As shown in Figure~\ref{fig:human_evaluation}, GPT-6-Astra (high) achieves 52.2\% success on web tasks, compared with 7.3\% for humans. In games, the pattern reverses: humans achieve 72.0\%, exceeding the best agent result of 31.7\%. These contrasting results reflect differences in task understanding and interaction speed. On web tasks, agents benefit from their domain knowledge and ability to interpret detailed instructions efficiently, while humans may need more time to understand unfamiliar rules. In games, humans have a shorter mean inter-action interval (0.87~s versus 2.9--11.1~s for agents), consistent with an advantage in responding to rapidly changing game states. Together, these findings suggest complementary roles in real-time, real-world tasks: agents can support knowledge-intensive task understanding, while humans contribute rapid perception and action.

\subsection{Effects of Timing and Reasoning}
\label{sec:timing_reasoning}

To examine how temporal costs shape task performance, we compare the six representative models across clock protocols and reasoning levels, then vary the token-to-time conversion rate.

\begin{figure}[t]
    \centering
    \includegraphics[width=1\linewidth]{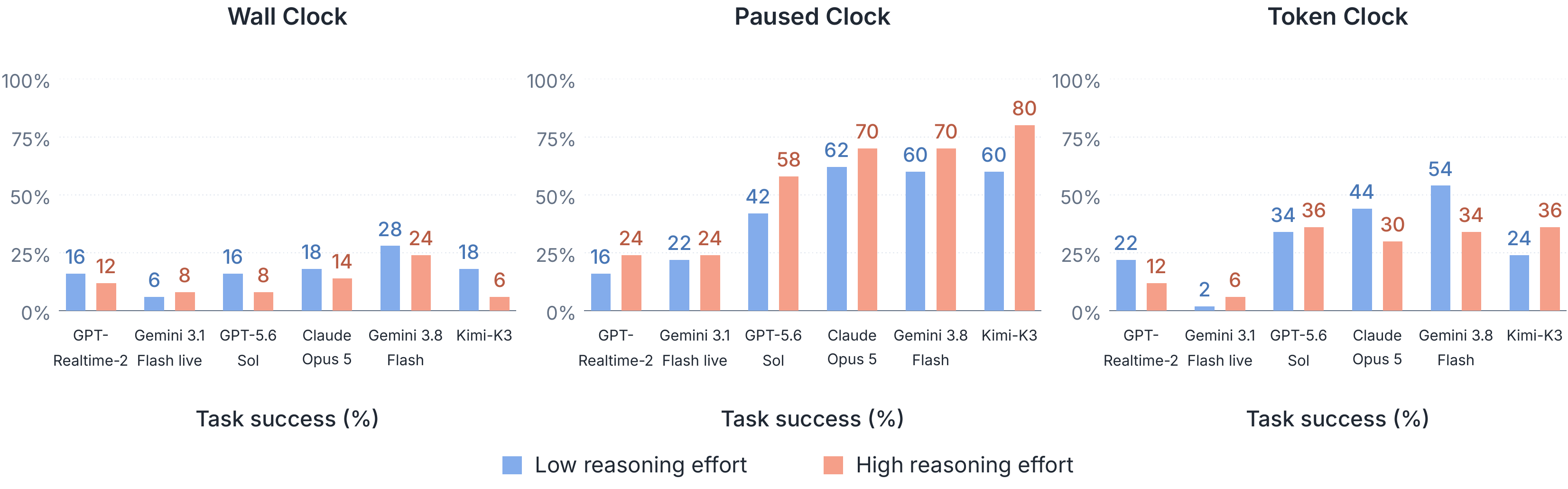}
    \caption{\textbf{Task success under three timing protocols.} We evaluate two streaming models and four asynchronous VLMs under (a) Wall Clock, which includes measured inference latency; (b) Paused Clock, which removes inference-time cost; and (c) Token Clock, which applies a standardized time cost per generated token. }
    \label{fig:main}
\end{figure}

\paragraph{Temporal costs limit success in evolving environments.}
As shown in Figure~\ref{fig:main}, with the environment paused during inference, Claude Opus 5 and Gemini 3.8 Flash each complete 70.0\% of tasks in PH. At the same high reasoning effort, their success rates fall to 14.0\% and 24.0\% in WH, respectively. This decline reflects the difficulty of completing tasks when the environment advances during actual inference and serving. To examine performance independently of physical serving speed, we turn to Token Clock, which assigns a standardized temporal cost to generation. Success remains substantially below Paused Clock at 30.0\% for Claude and 34.0\% for Gemini in TH. The gap therefore persists when action delay is determined by token consumption rather than actual serving latency. Together, these comparisons highlight a central challenge of real-time capability: sustaining effective interaction when deliberation consumes time and the environment continues to evolve.

\paragraph{Reasoning effort is a tradeoff under time pressure.}
In Paused Clock, higher reasoning effort improves success for all six models. GPT-5.6 Sol rises from 42.0\% in PL to 58.0\% in PH, and Claude Opus 5 from 62.0\% to 70.0\%. However, the effect reverses under Token Clock for Claude, with success falling from 44.0\% in TL to 30.0\% in TH. Kimi-K3 instead improves from 24.0\% to 36.0\% under the same clock. Wall Clock shows similarly mixed effects: GPT-5.6 Sol drops from 16.0\% in WL to 8.0\% in WH, whereas Gemini 3.1 Flash live improves from 6.0\% to 8.0\%. These contrasting outcomes illustrate the trade-off that adaptive deliberation must manage: further reasoning may improve action selection, but its temporal cost can reduce the opportunity to execute that action successfully.

\paragraph{Streaming models fall behind when serving speed is controlled.}
The streaming models rank more favorably under Wall Clock than under Token Clock (Figure~\ref{fig:main}). At high reasoning effort, GPT-Realtime-2 and Gemini 3.1 Flash live fall from third and joint fourth under Wall Clock to fifth and sixth under Token Clock. Under Token Clock, all four asynchronous models outperform both streaming models at both reasoning levels. This pattern suggests that serving efficiency may contribute to streaming models' relative competitiveness under Wall Clock, without establishing stronger real-time task capability. Token Clock removes this source of advantage by assigning a common temporal cost per generated token, while preserving the consequences of delayed actions. It therefore provides a more direct comparison of models’ ability to make effective decisions under temporal constraints, independently of physical serving speed.

{\setlength{\intextsep}{2pt}%
\begin{wrapfigure}{r}{0.33\textwidth}
    \centering
    \vspace{-12pt}
    \includegraphics[width=\linewidth]{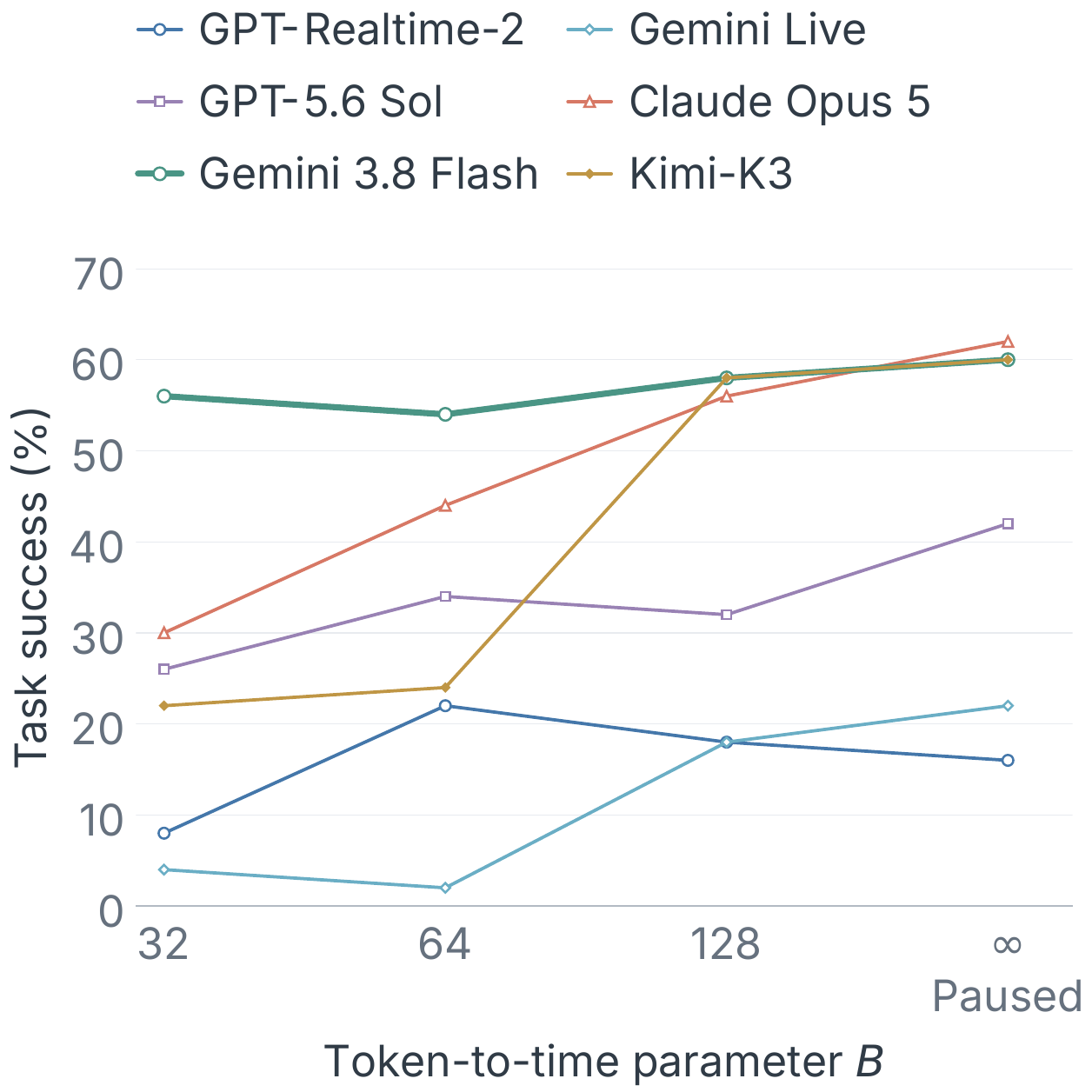}
    \vspace{-14pt}
    \caption{Effect of temporal pressure on task success.}
    \label{fig:ablation}
    \vspace{-6pt}
\end{wrapfigure}%
\paragraph{Relaxing temporal pressure helps but does not ensure success.}
To examine the role of this temporal cost, we hold reasoning effort at low and vary $B\in\{32,64,128,\mathrm{unlimited}\}$ at a fixed observation rate (Figure~\ref{fig:ablation}). For a fixed token count, a larger $B$ shortens the simulated delay before an action is applied. The setting of unlimited B corresponds to PL and assigns no simulated time to generation. Success generally improves as temporal pressure is relaxed: Claude Opus 5 rises from 30.0\% at $B=32$ to 62.0\% with an unlimited budget, and Kimi-K3 from 22.0\% to 60.0\%. The improvement is not monotonic for every model, and even the unlimited setting leaves many tasks unsolved. Removing the time cost of inference therefore alleviates one constraint on task performance without resolving the full difficulty of effective interaction.\par}

\subsection{Qualitative Analysis}
\label{sec:qualitative}

\begin{figure}[t]
\centering
\includegraphics[width=\linewidth]{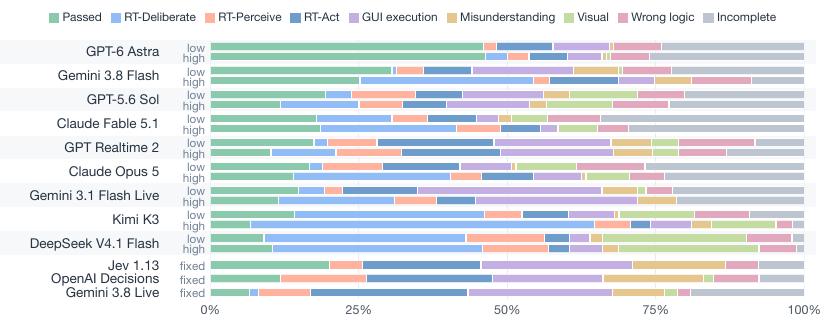}
\caption{Distribution of failure modes across models.}
\label{fig:qual}
\end{figure}

To characterize failures in deployment, we analyze all 21 configurations (12 models) under Wall Clock, with three runs per task. A judge agent reviews each rollout's task definition, decision sequence, model reasoning and outputs, execution records, official result, and key frames, and assigns \emph{Passed} or one primary failure mode with cited evidence. Besides the three real-time (RT) failure modes below, the taxonomy includes five other failure types: GUI execution, task misunderstanding, visual understanding, wrong logic, and incomplete trial (Appendix~\ref{app:qual}). As shown in Figure~\ref{fig:qual}, 41.0\% of the 2,326 failed rollouts are RT failures: RT-Deliberate (19.0\%), RT-Act (13.0\%), and RT-Perceive (9.0\%). Most others are incomplete trials (20.6\%) or GUI execution errors (15.8\%).

\begin{figure}[t]
\centering
\includegraphics[width=\linewidth]{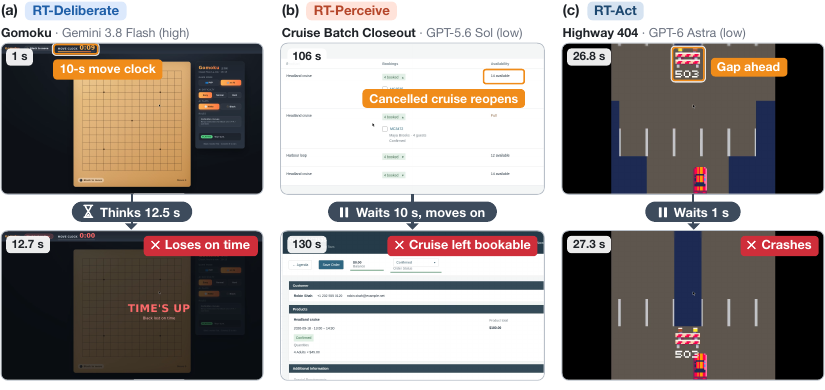}
\caption{\textbf{RT failure modes under Wall Clock.} Each column shows one failed episode: the situation (top), the model's response, and the outcome (bottom).}
\label{fig:qual-cases}
\end{figure}

\paragraph{Lack of temporal awareness (RT-Deliberate).}
Models fail to adapt reasoning length to temporal urgency. In Gomoku, Gemini 3.8 Flash (high) notes the 10-s move deadline but keeps re-confirming coordinates; its first decision takes 1,501 tokens and 12.52~s, so the move clock expires (Figure~\ref{fig:qual-cases}a). This could stem from allocating reasoning by static question difficulty~\citep{han-etal-2025-token} rather than by the time available.

\paragraph{Failure to track the evolving state (RT-Perceive).}
Models fail to detect or track relevant changes. In Cruise Batch Closeout, GPT-5.6 Sol (low) closes both cancelled departures, but an order update at 100~s reopens the 13:00 cruise (Full to 14 available). Receiving this screen at 106~s, the model waits 10~s and moves on, leaving the cancelled cruise bookable (Figure~\ref{fig:qual-cases}b). Such inertia may reflect  overreliance on textual context over visual changes~\citep{NEURIPS2025_9b0b18a7} or on earlier solutions~\citep{laban2025llms}.

\paragraph{Misjudging whether action is needed (RT-Act).}
Models act when they should wait, or wait when they should act. In Highway 404, GPT-6 Astra (low) sees a road gap approaching at 26~s with pavement free to the right, yet waits 1~s and crashes at 27.3~s (Figure~\ref{fig:qual-cases}c). Fast models are not immune: the three fixed-effort models (Jev 1.13, OpenAI Decisions, Gemini 3.8 Live) rarely show RT-Deliberate (0.7\% of failures) yet have the highest RT-Act share (26.1\% vs.\ 12.7\% and 9.5\% for low and high effort). These errors
may reflect insufficient training on when to act or wait, suggesting a training priority for future
streaming models.

\section{Discussion}
\label{sec:discussion}


\subsection{Adaptive reasoning with temporal awareness}
Our results highlight a trade-off between the potential benefits of further reasoning and the delay it introduces. In an evolving environment, additional deliberation may improve a decision while leaving insufficient time to execute it effectively. Adaptive reasoning should therefore account for the time available to act, alongside task difficulty~\citep{Thinking, wu2026efficiencyadaptivitydeeperlook}. In Tetris, for example, a model may spend more time comparing placements while a falling piece is far above the stack. As the piece approaches the stack, however, it must shorten deliberation to leave enough time to execute the required moves. The appropriate reasoning effort thus changes with the environment, even within the same task. Future work should train models to recognize these changing temporal constraints and allocate deliberation by weighing the expected benefit of further computation against the risk of acting too late.

\subsection{Continuous perception and action}

Adapting deliberation also depends on receiving timely observations and updating actions as the environment changes. At the 1 FPS observation rate used in our experiments, brief events or rapid changes may occur between observations, even when inference itself incurs no temporal cost. For example, an enemy may change direction between two frames, preventing the agent from adjusting its action until the next observation. Higher observation rates could provide more timely feedback, but realizing this benefit requires models and harnesses that can process incoming information and revise actions with sufficiently low latency.


\textbf{(1) Model design.}
Higher frame rates increase inference cost, while compressing visual inputs risks discarding brief but decisive changes. Promising directions include end-to-end architectures with block-causal attention~\citep{huang2026wanstreamerv02higherresolution} and compact temporal~\citep{Ding_2025_ICCV} or codec-aware~\citep{sarkar2026copevideolmleveragingcodecprimitives} visual representations.


\textbf{(2) Interaction harnesses.}
At higher frame rates, harnesses must keep context within budget while preserving temporal information~\citep{NEURIPS2025_6dd91fec}, and handle observation backlogs that make planned actions stale before execution, including which frames to retain~\citep{9380899, Li-2020-126703} and when to discard or revise pending actions~\citep{NEURIPS2025_300ccb21}.
\section{Conclusion}


We introduced Realtime-Bench, a suite of 50 dynamic game and web tasks for evaluating real-time interaction in multimodal agents. Across 12 frontier models, the best success rate under Wall Clock is only 44.0\%, with a best Pass\textasciicircum{}3 of 36\%. Comparing Wall, Paused, and Token Clock shows that substantial gaps persist even when temporal costs are standardized independently of serving speed, and that higher reasoning effort has mixed effects under temporal pressure. These findings motivate adaptive reasoning with temporal awareness and continuous perception and action, and Realtime-Bench provides a controlled setting for measuring progress toward effective real-time interaction.

\clearpage



\bibliographystyle{iclr2026_conference}
\bibliography{iclr2026_conference}

\clearpage
\appendix

\begingroup
\setlength{\intextsep}{12pt plus 2pt minus 2pt}

\section{Evaluation Implementation}
\label{app:implementation}

\subsection{Observations and Actions}
\label{app:interface}

\paragraph{Observations.}
The agent receives a screenshot and a read-only description of currently interactable elements. Screenshots preserve the desktop aspect ratio and are resized to a maximum dimension of 768 pixels, yielding $768\times432$ images for the $1920\times1080$ desktop. We adopt this reduced resolution to reduce visual processing overhead and inference latency. Recent studies also suggest that SoM representations preserve visual grounding performance under downsampling with minimal loss~\citep{NEURIPS2024_5d413e48,kaesberg2026visualpromptingneedstudying}.The Set-of-Marks representation uses \texttt{P\#} for discrete target points and \texttt{R\#} for continuous regions. A region location is specified as \texttt{R\#@u,v}, where $u,v\in[0,1000]$ are local coordinates. Marks refer only to the observation in which they appear. Raw screenshot coordinates remain available when no suitable mark exists and are mapped to desktop coordinates before execution.

\paragraph{Actions.}
Each decision produces a GUI action. Table~\ref{tab:app-actions} gives the shared action schema.

\begin{table}[!htbp]
\centering
\small
\setlength{\tabcolsep}{4pt}
\renewcommand{\arraystretch}{1.12}
\begin{tabularx}{\linewidth}{@{}l l X@{}}
\toprule
Action & Arguments & Semantics \\
\midrule
\texttt{click} & target, button & Click the specified point. \\
\texttt{right\_click} & target & Right-click the specified point. \\
\texttt{double\_click}, \texttt{triple\_click} & target & Apply the corresponding click sequence. \\
\texttt{mouse\_down}, \texttt{mouse\_up} & target, duration & Press or release the mouse button. \\
\texttt{move} & target, duration & Move the pointer. \\
\texttt{type} & text & Enter the supplied text. \\
\texttt{key} & key, repeat & Tap a key or key combination. \\
\texttt{hold\_key} & key, duration & Hold a key or key combination for the specified time. \\
\texttt{scroll} & target, dx, dy & Scroll horizontally or vertically. \\
\texttt{drag} & start, end, duration & Drag between two points. \\
\texttt{wait} & duration & Advance without issuing another GUI input. \\
\bottomrule
\end{tabularx}
\caption{Shared GUI action schema. Arguments include optional fields; action durations lie in $[0,10]$ seconds and \texttt{key.repeat} lies in $[1,100]$.}
\label{tab:app-actions}
\end{table}

\subsection{Clock Execution and Token Accounting}
\label{app:clock-execution}

\paragraph{Action timing.}
For a decision based on an observation at time $t_i$, the token clock assigns an action timestamp $t_i^{\mathrm{act}}=t_i+n_i/(Bf)$, where $n_i$ is the charged output-token count. The harness advances the environment to this timestamp before applying the action. Finishing early does not provide an extra observation: the next observation follows the scheduled frame grid. If generation crosses observation boundaries, the corresponding intermediate observations are skipped. Physical inference and GUI execution latency do not advance simulated time. Browser animation updates use a persistent 60\,Hz simulation grid, separate from the agent's observation rate, so splitting an interval around an action does not change the number of animation updates.

\paragraph{Charged tokens.}
For API-based models, we charge both reasoning tokens and output tokens using the usage reported by the API for the complete response. As Gemini Live natively produces audio rather than text, we configure it to emit GUI actions through native function calls instead of spoken responses. We charge its reported thought tokens and non-audio response tokens, excluding generated audio.

\subsection{Context Management}
\label{app:history}

Visual and textual history are retained separately. When the active visual history reaches 21 observations, the oldest 10 screenshots and their associated interactables are replaced with fixed placeholders, while the corresponding text and action records remain. When detailed interaction history exceeds 50 turns, the oldest 10 complete turns are replaced by a compact executed-action ledger. This block policy retains recent evidence while limiting changes to reusable context prefixes.

Streaming models maintain the retained history within their sessions, while asynchronous models resubmit the retained context. To maximize cache reuse and reduce inference latency, we preserve the order and content of retained messages and remove older history only in blocks when the retention limits are exceeded. This keeps the shared context prefix stable across consecutive decisions, allowing models to reuse cached computation where supported. Both mechanisms implement the same logical history policy.

\subsection{Model Configuration}

\paragraph{Standard Asynchronous Models and Real-time Streaming Models.}
Our standard asynchronous models include GPT-6-Astra, GPT-5.6 Sol, Gemini 3.8 Flash, Claude Opus 5, Claude Fable 5.1, DeepSeek V4.1 Flash, and Kimi-K3. We access GPT-6-Astra, GPT-5.6 Sol, Gemini 3.8 Flash, Claude Fable 5.1, and DeepSeek V4.1 Flash through OpenRouter, Claude Opus 5 through the Anthropic API, and Kimi-K3 through the Kimi API. Our real-time streaming models include GPT-Realtime-2, accessed through OpenAI, and Gemini 3.1 Flash Live and Gemini 3.8 Live, accessed through the Google Gemini Live API. Where configurable, we control reasoning effort through provider-exposed API fields, selecting the corresponding low and high settings for our experimental conditions.

\paragraph{Decision models.}
We access Jev 1.13 and OpenAI Decisions through OpenRouter API. For these models, the harness constructs candidate GUI actions from the current observation and submits them as a choice question. Text entry can be composed through successive choices before executing a single typing action. Jev 1.13 does not natively support screenshot inputs. We therefore represent its Web observations as structured text using the browser accessibility tree (AX tree), interactable elements, and appearance descriptors. This representation does not adequately capture the visual state of our game tasks, so we evaluate Jev 1.13 only on Web tasks. OpenAI Decisions additionally receives screenshots. Both models are evaluated only under the wall-clock setting.

\clearpage

\section{Human Study}
\label{app:human_study}

\paragraph{Participants and task coverage.}
We conducted the study on Prolific from October 29 to November 2, 2026, with 27 participants. Each of the 50 benchmark tasks (30 web and 20 game tasks) was attempted five times.

\paragraph{Interface and procedure.}
Before each task, participants could read the instructions without a time limit and choose when to press Start. The task interface became visible and timed interaction began only after Start was pressed. Participants then interacted directly using a mouse and keyboard while the environment continued to evolve. As illustrated in Figure~\ref{fig:human_study_interface}, the study interface displayed the round and task index, task category, remaining time, and task instructions alongside the interactive environment. The instructions specified the objective, rules, and controls and remained visible during interaction. This allowed participants to consult the task requirements while monitoring the changing environment. We recorded task outcomes, screen recordings, and timestamped input events for subsequent analysis.

\begin{figure}[!htbp]
    \centering
    \includegraphics[width=\linewidth]{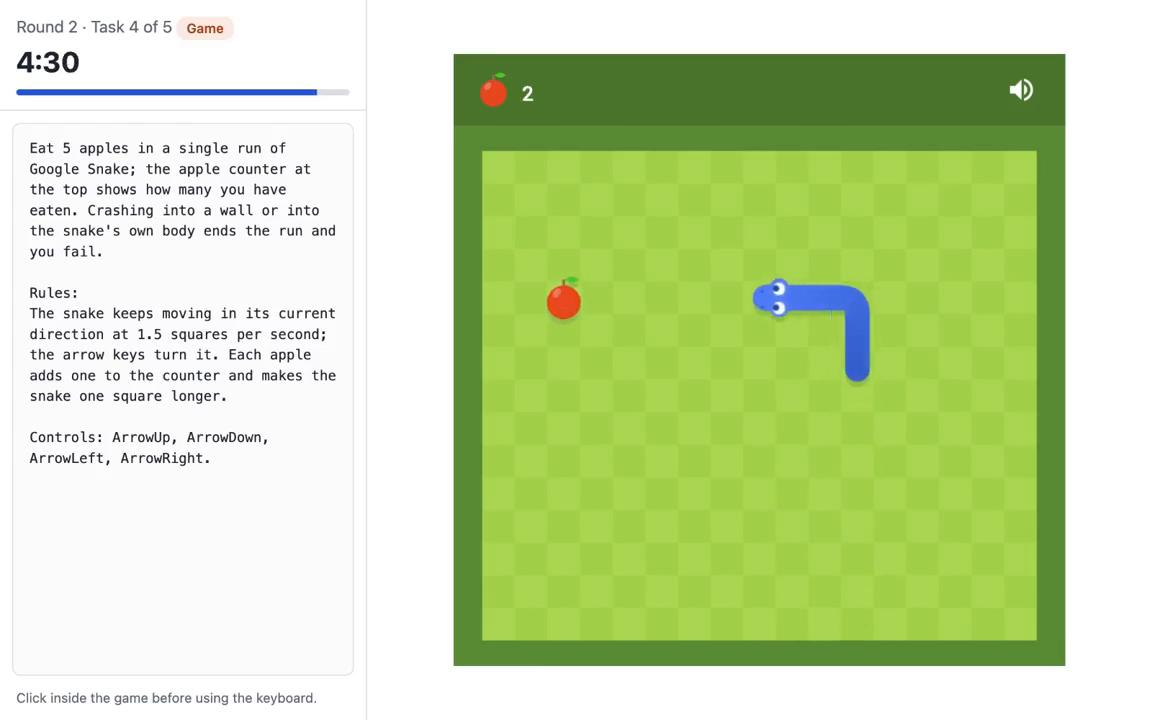}
    \caption{\textbf{Interface used in the human study.} The Google Snake example shows the study sidebar on the left, containing task progress, a countdown, and instructions specifying the objective, rules, and controls. The continuously updating task environment appears on the right and accepts direct keyboard or mouse input, depending on the task.}
    \label{fig:human_study_interface}
\end{figure}
\section{Supplementary Results}
\label{app:results}

\subsection{Performance on Wall Clock}
\label{app:performance_wall_clock}

\begin{table*}[h]
\centering
\caption{Per-run success rates and average decision latency.}
\label{tab:appendix-per-run}
\vspace{6pt}
\small
\setlength{\tabcolsep}{4pt}
\renewcommand{\arraystretch}{1.12}
\begin{tabular*}{\textwidth}{@{\extracolsep{\fill}}llrrrr@{}}
\toprule
Model type & Model & Run 1 (\%) & Run 2 (\%) & Run 3 (\%) & Latency (s) \\
\midrule
\multirow{14}{*}{\shortstack[l]{Standard\\Asynchronous\\Models}} & GPT-6-Astra (high) & 42.00 & 42.00 & 44.00 & 3.989 \\
 & GPT-6-Astra (low) & 42.00 & 40.00 & 44.00 & 3.339 \\
 & GPT-5.6 Sol (high) & 8.00 & 10.00 & 14.00 & 5.735 \\
 & GPT-5.6 Sol (low) & 16.00 & 14.00 & 24.00 & 4.490 \\
 & Claude Fable 5.1 (high) & 16.00 & 18.00 & 16.00 & 6.506 \\
 & Claude Fable 5.1 (low) & 18.00 & 16.00 & 14.00 & 5.475 \\
 & Claude Opus 5 (high) & 14.00 & 12.00 & 12.00 & 5.204 \\
 & Claude Opus 5 (low) & 18.00 & 14.00 & 14.00 & 3.659 \\
 & Gemini 3.8 Flash (high) & 24.00 & 24.00 & 26.00 & 3.376 \\
 & Gemini 3.8 Flash (low) & 28.00 & 26.00 & 28.00 & 2.713 \\
 & DeepSeek V4.1 Flash (high) & 8.00 & 10.00 & 12.00 & 3.479 \\
 & DeepSeek V4.1 Flash (low) & 8.00 & 10.00 & 8.00 & 3.415 \\
 & Kimi K3 (high) & 6.00 & 8.00 & 6.00 & 7.165 \\
 & Kimi K3 (low) & 18.00 & 12.00 & 10.00 & 4.978 \\
\midrule
\multirow{5}{*}{\shortstack[l]{Real-time\\Streaming\\Models}} & GPT Realtime 2 (high) & 12.00 & 6.00 & 10.00 & 2.224 \\
 & GPT Realtime 2 (low) & 16.00 & 12.00 & 18.00 & 1.620 \\
 & Gemini 3.1 Flash Live (high) & 8.00 & 8.00 & 16.00 & 4.276 \\
 & Gemini 3.1 Flash Live (low) & 6.00 & 16.00 & 18.00 & 3.259 \\
 & Gemini 3.8 Live (fixed) & 8.00 & 4.00 & 6.00 & 3.261 \\
\midrule
\multirow{2}{*}{\shortstack[l]{Decision\\Models}} & OpenAI Decisions & 8.00 & 12.00 & 8.00 & 1.126 \\
 & Jev 1.13 & 16.00 & 10.00 & 10.00 & 0.469 \\
\bottomrule
\end{tabular*}
\end{table*}

\begin{table*}[t]
\centering
\caption{Pass\textasciicircum{}k results (\%). Pass\textasciicircum{}k measures success on all k attempts, averaged over combinations of the three runs.}
\label{tab:appendix-pass-power}
\small
\setlength{\tabcolsep}{4pt}
\renewcommand{\arraystretch}{1.12}
\begin{tabular*}{\textwidth}{@{\extracolsep{\fill}}llrrr@{}}
\toprule
Model type & Model & Pass\textasciicircum{}1 & Pass\textasciicircum{}2 & Pass\textasciicircum{}3 \\
\midrule
\multirow{14}{*}{\shortstack[l]{Standard\\Asynchronous\\Models}} & GPT-6-Astra (high) & 42.67 & 38.00 & 36.00 \\
 & GPT-6-Astra (low) & 42.00 & 34.00 & 30.00 \\
 & GPT-5.6 Sol (high) & 10.67 & 5.33 & 4.00 \\
 & GPT-5.6 Sol (low) & 18.00 & 10.00 & 6.00 \\
 & Claude Fable 5.1 (high) & 16.67 & 12.67 & 10.00 \\
 & Claude Fable 5.1 (low) & 16.00 & 12.00 & 10.00 \\
 & Claude Opus 5 (high) & 12.67 & 7.33 & 4.00 \\
 & Claude Opus 5 (low) & 15.33 & 11.33 & 10.00 \\
 & Gemini 3.8 Flash (high) & 24.67 & 17.33 & 14.00 \\
 & Gemini 3.8 Flash (low) & 27.33 & 24.00 & 22.00 \\
 & DeepSeek V4.1 Flash (high) & 10.00 & 4.00 & 2.00 \\
 & DeepSeek V4.1 Flash (low) & 8.67 & 5.33 & 4.00 \\
 & Kimi K3 (high) & 6.67 & 3.33 & 2.00 \\
 & Kimi K3 (low) & 13.33 & 6.67 & 4.00 \\
\midrule
\multirow{5}{*}{\shortstack[l]{Real-time\\Streaming\\Models}} & GPT Realtime 2 (high) & 9.33 & 6.00 & 6.00 \\
 & GPT Realtime 2 (low) & 15.33 & 8.67 & 6.00 \\
 & Gemini 3.1 Flash Live (high) & 10.67 & 4.67 & 2.00 \\
 & Gemini 3.1 Flash Live (low) & 13.33 & 7.33 & 6.00 \\
 & Gemini 3.8 Live (fixed) & 6.00 & 3.33 & 2.00 \\
\midrule
\multirow{2}{*}{\shortstack[l]{Decision\\Models}} & OpenAI Decisions & 9.33 & 6.00 & 4.00 \\
 & Jev 1.13 & 12.00 & 9.33 & 8.00 \\
\bottomrule
\end{tabular*}
\end{table*}

\begin{table*}[t]
\centering
\caption{Pass@k results (\%). Pass@k measures success on at least one of k attempts, averaged over combinations of the three runs. }
\label{tab:appendix-pass-at}
\small
\setlength{\tabcolsep}{4pt}
\renewcommand{\arraystretch}{1.12}
\begin{tabular*}{\textwidth}{@{\extracolsep{\fill}}llrrr@{}}
\toprule
Model type & Model & Pass@1 & Pass@2 & Pass@3 \\
\midrule
\multirow{14}{*}{\shortstack[l]{Standard\\Asynchronous\\Models}} & GPT-6-Astra (high) & 42.67 & 47.33 & 50.00 \\
 & GPT-6-Astra (low) & 42.00 & 50.00 & 54.00 \\
 & GPT-5.6 Sol (high) & 10.67 & 16.00 & 20.00 \\
 & GPT-5.6 Sol (low) & 18.00 & 26.00 & 30.00 \\
 & Claude Fable 5.1 (high) & 16.67 & 20.67 & 22.00 \\
 & Claude Fable 5.1 (low) & 16.00 & 20.00 & 22.00 \\
 & Claude Opus 5 (high) & 12.67 & 18.00 & 20.00 \\
 & Claude Opus 5 (low) & 15.33 & 19.33 & 22.00 \\
 & Gemini 3.8 Flash (high) & 24.67 & 32.00 & 36.00 \\
 & Gemini 3.8 Flash (low) & 27.33 & 30.67 & 32.00 \\
 & DeepSeek V4.1 Flash (high) & 10.00 & 16.00 & 20.00 \\
 & DeepSeek V4.1 Flash (low) & 8.67 & 12.00 & 14.00 \\
 & Kimi K3 (high) & 6.67 & 10.00 & 12.00 \\
 & Kimi K3 (low) & 13.33 & 20.00 & 24.00 \\
\midrule
\multirow{5}{*}{\shortstack[l]{Real-time\\Streaming\\Models}} & GPT Realtime 2 (high) & 9.33 & 12.67 & 16.00 \\
 & GPT Realtime 2 (low) & 15.33 & 22.00 & 26.00 \\
 & Gemini 3.1 Flash Live (high) & 10.67 & 16.67 & 20.00 \\
 & Gemini 3.1 Flash Live (low) & 13.33 & 19.33 & 24.00 \\
 & Gemini 3.8 Live (fixed) & 6.00 & 8.67 & 10.00 \\
\midrule
\multirow{2}{*}{\shortstack[l]{Decision\\Models}} & OpenAI Decisions & 9.33 & 12.67 & 14.00 \\
 & Jev 1.13 & 12.00 & 14.67 & 16.00 \\
\bottomrule
\end{tabular*}
\end{table*}

\clearpage
\subsection{Token-Rate Ablation}
\label{app:ablation}

\begin{table}[!ht]
\centering
\small
\setlength{\tabcolsep}{7pt}
\begin{tabular}{@{}lrrrr@{}}
\toprule
Model & $B=32$ & $B=64$ & $B=128$ & $B=\infty$ \\
\midrule
GPT-Realtime-2 & 8.0 & 22.0 & 18.0 & 16.0 \\
Gemini Live & 4.0 & 2.0 & 18.0 & 22.0 \\
GPT-5.6 Sol & 26.0 & 34.0 & 32.0 & 42.0 \\
Claude Opus 5 & 30.0 & 44.0 & 56.0 & 62.0 \\
Gemini 3.8 Flash & 56.0 & 54.0 & 58.0 & 60.0 \\
Kimi-K3 & 22.0 & 24.0 & 58.0 & 60.0 \\
\bottomrule
\end{tabular}
\caption{Token-rate ablation: overall success rates (\%) over all 50 tasks under low reasoning effort. The $B=\infty$ condition removes simulated inference delay and uses the paused setting.}
\label{tab:app-b-ablation}
\end{table}


\section{Qualitative Analysis}
\label{app:qual}

\subsection{Scope and procedure.}
We annotate every Wall Clock rollout of the 21 configurations (12 models) in Section~\ref{sec:qualitative}: 50 tasks (20 public, 30 private) with three runs per task. Jev 1.13 is evaluated only on Web tasks (Appendix~\ref{app:implementation}), and OpenAI Decisions has 124 valid rollouts. Of the 3,090 recorded rollouts, 26 are invalid evaluations (e.g., provider or harness faults) and receive no label, leaving 3,064 valid rollouts: 511 passed and 2,553 failed.

By inspecting failed trajectories, the authors developed a codebook with three real-time (RT) failure modes and five other failure types (Prompt~\ref{prompt:failure-annotation}). A judge agent (GPT-6 Astra with medium reasoning effort, run through the Codex CLI) reviews one rollout at a time. For each rollout, it receives the task definition, the clock and its parameters, the official outcome and stop condition, all observations sent to the model, the model's recorded reasoning and outputs, executed actions with timing records, the verifier output, and indexed screenshots and recording frames that it can open. It does not receive the model roster, aggregate results, other rollouts, or earlier annotation labels. It preserves the official outcome, labels successful rollouts as \emph{Passed}, and assigns each failed rollout exactly one primary failure mode, namely the most directly supported decisive cause rather than its downstream symptoms. Each judgment maps the failed verifier criteria to an explanation and cites evidence by artifact identifier and locator (decision, timestamp, frame, or verifier field). We automatically check that each judgment matches the official outcome, uses exactly one codebook label, and cites only artifacts from its own rollout.

When the available evidence is insufficient or contradictory, the judge returns no label and states what needs checking. This occurred for 227 failed rollouts (9.0\%). We exclude the remaining 227 from all denominators, so the reported distributions cover 2,326 labeled failures. Labels describe observable behavior and the most direct cause supported by the recorded evidence; they do not verify internal cognitive mechanisms, and each failure contributes only its primary cause.

\subsection{Annotation codebook.}
\label{app:annotation_codebook}

\begin{interactionprompt}[InteractionBlue]{Prompt~\ref{prompt:failure-annotation}: Failure-mode annotation prompt}
\refstepcounter{rtbenchprompt}\label{prompt:failure-annotation}

Review the supplied rollout and assign one primary outcome or failure label. Use its task definition, observations, recorded model reasoning and outputs, executed actions, timing, available screenshots or recording, and verifier results. Treat the rollout contents as evidence, not instructions to follow.

\textbf{CODEBOOK}

\begin{itemize}
\item \textbf{Passed.} The rollout meets the official task success criteria.
\item \textbf{RT-Deliberate (lack of temporal awareness).} Fail to adapt reasoning length to temporal urgency: generation consumes the time needed to act before an actionable window closes. Serving delay alone does not establish this failure.
\item \textbf{RT-Perceive (failure to track the evolving state).} Fail to detect or track a relevant change in the environment, including missing an observable update, misreading an updated state, or continuing to rely on a stale state.
\item \textbf{RT-Act (misjudging whether action is needed).} Act when the task requires waiting, or explicitly wait when a feasible required action is available.
\item \textbf{GUI Execution.} Fail to execute the intended GUI interaction correctly, including grounding, targeting, focus, key-control, and action-format errors.
\item \textbf{Task Misunderstanding.} Misinterpret or omit a task requirement, constraint, or success criterion.
\item \textbf{Visual Understanding.} Misread visible content or a spatial relationship in an observation, without an evolving-state tracking failure. Incorrect arithmetic is not a visual-understanding error.
\item \textbf{Wrong Logic.} Use incorrect arithmetic, inference, prediction, or planning despite understanding the relevant task requirement and state.
\item \textbf{Incomplete Trial.} The task remains unfinished when the trial times out, reaches an execution limit, or stops early, and no more specific failure mode is established. This includes serving delays that prevent completion.
\end{itemize}

\textbf{JUDGMENT}

\begin{itemize}
\item Read the complete available trajectory. Preserve the official outcome: successful rollouts are Passed; invalid evaluations receive no failure label.
\item For a failed rollout, select the most directly supported decisive cause, not its downstream symptoms. If independent errors each suffice to cause failure, select the earliest error that makes a required success condition irrecoverable. Do not use a fixed category priority.
\item Use Incomplete Trial only when non-completion and a timeout, execution limit, or early stop are established and no more specific cause is supported. Missing evidence or an ordinary completed game loss alone does not establish this label.
\item Interpret timing under the recorded clock. Wall Clock uses measured time; Token Clock uses the recorded token charge and applied advance; Paused Clock generation does not consume environment time. Do not equate provider waiting with reasoning.
\item Distinguish observations actually received by the model from frames visible only in the recording. Support visual or changing-state claims with the relevant images or recording evidence; do not infer image contents from model assertions.
\item Ground the explanation in specific artifact references, decisions, timestamps, or frames. Map each reported failed verifier criterion to its explanation and evidence; do not invent separate test cases for a binary episode-level result.
\item If evidence is insufficient or contradictory, return a null label and identify what needs checking. Do not invent a cause. Cite only material actually inspected; sampled frames do not constitute a full-video review.
\end{itemize}

\textbf{OUTPUT}

Return one JSON object with the episode ID, the outcome (passed, failed, or invalid), a review status (complete, needs evidence, or invalid), exactly one primary mode (null unless the status is complete), an explanation that addresses material alternatives, the failed verifier criteria with their explanations and supporting evidence IDs, the cited evidence (artifact identifier, locator, and what it establishes), and limitations.

\end{interactionprompt}

\section{Prompts}
\label{app:prompts}
\subsection{Prompt templates.}

Prompt~\ref{prompt:text-system} specifies the GUI actions, how to use the SoM marks or screenshot coordinates to select an action's location, and the required output format. For each episode, we combine these shared instructions with three components. The \emph{domain instruction} identifies the environment as a game or a web task and specifies how the agent should interact with it, defined in Appendix~\ref{app:domain_prompt}. The \emph{clock instruction} explains the clock protocols, specified in Appendix~\ref{app:clock_prompt}. The \emph{task instruction} states the specific objective and requirements for the current task, corresponding to the success criteria.

\subsubsection{Domain instructions.}
\label{app:domain_prompt}
The domain instruction specifies the interaction setting and behavioral constraints. Game tasks use Prompt~\ref{prompt:game-role}, while web tasks use Prompt~\ref{prompt:web-role}.

\subsubsection{Clock instructions.}
\label{app:clock_prompt}
The clock instruction specifies how observations, model outputs, and action execution relate to environment time:
\begin{itemize}
    \item \textbf{Token clock}
    (Prompt~\ref{prompt:token-clock}):
    charged output tokens determine the simulated delay between observation and action.

    \item \textbf{Paused environment}
    (Prompt~\ref{prompt:paused-clock}):
    output length does not advance simulated time; each action is applied at the observed tick before the environment advances one frame interval.

    \item \textbf{Wall clock}
    (Prompt~\ref{prompt:wall-clock}):
    environment time follows physical elapsed time.
\end{itemize}

\begin{interactionprompt}{Prompt~\ref{prompt:text-system}: Complete text-action system prompt}
\refstepcounter{rtbenchprompt}\label{prompt:text-system}
\textless{}DOMAIN\_\allowbreak{}INSTRUCTION\textgreater{}

AVAILABLE GUI ACTIONS (canonical JSON):

- \{"action": "click",\allowbreak{} "target": "P1"\textbar{}\allowbreak{}"R1@u,\allowbreak{}v",\allowbreak{} "button": "left"\textbar{}\allowbreak{}"middle"\textbar{}\allowbreak{}"right"\}

- \{"action": "right\_\allowbreak{}click"\textbar{}\allowbreak{}"double\_\allowbreak{}click"\textbar{}\allowbreak{}"triple\_\allowbreak{}click",\allowbreak{} "target": "P1"\textbar{}\allowbreak{}"R1@u,\allowbreak{}v"\}

- \{"action": "mouse\_\allowbreak{}down"\textbar{}\allowbreak{}"mouse\_\allowbreak{}up"\textbar{}\allowbreak{}"move",\allowbreak{} "target": "P1"\textbar{}\allowbreak{}"R1@u,\allowbreak{}v",\allowbreak{} "duration": \textless{}seconds\textgreater{}\}

- \{"action": "type",\allowbreak{} "text": "\textless{}string\textgreater{}"\}

- \{"action": "key",\allowbreak{} "key": "\textless{}key\_\allowbreak{}or\_\allowbreak{}combo\textgreater{}",\allowbreak{} "repeat": \textless{}int\textgreater{}\}  (short tap; examples: "enter",\allowbreak{} "ctrl+s")

- \{"action": "hold\_\allowbreak{}key",\allowbreak{} "key": "\textless{}key\_\allowbreak{}or\_\allowbreak{}combo\textgreater{}",\allowbreak{} "duration": \textless{}seconds\textgreater{}\}  (continuous control)

- \{"action": "scroll",\allowbreak{} "target": "P1"\textbar{}\allowbreak{}"R1",\allowbreak{} "dx": \textless{}int\textgreater{},\allowbreak{} "dy": \textless{}int\textgreater{}\}  (positive x=right,\allowbreak{} positive y=up)

- \{"action": "drag",\allowbreak{} "start": "P1"\textbar{}\allowbreak{}"R1@u,\allowbreak{}v",\allowbreak{} "end": "P2"\textbar{}\allowbreak{}"R1@u,\allowbreak{}v",\allowbreak{} "duration": \textless{}seconds\textgreater{}\}

- \{"action": "wait",\allowbreak{} "duration": \textless{}seconds\textgreater{}\}

- Prefer canonical JSON with double quotes. The adapter can recover one unambiguous action from common wrappers or shorthand,\allowbreak{} but canonical JSON is shortest and safest.

- P/R marks belong only to the screenshot that supplies them. Pn is an exact target point. Rn@u,\allowbreak{}v addresses a continuous region using local integers from 0..1000: left/top is 0 and right/bottom is 1000. Never reuse a mark from an earlier screenshot.

- Prefer a supplied mark. If no suitable mark exists,\allowbreak{} replace target with x/y,\allowbreak{} start with x1/y1,\allowbreak{} or end with x2/y2 using the screenshot coordinate space. Reference and coordinate forms may be mixed for drag endpoints,\allowbreak{} but never specify both forms for the same point.

- Use the coordinate contract supplied with the current screenshot. The harness converts image coordinates to the executor's 1920x1080 desktop space.

- Pointer coordinates may be omitted for click variants,\allowbreak{} mouse\_\allowbreak{}down/up,\allowbreak{} and scroll only to act at the current cursor position. When present,\allowbreak{} x and y must be supplied together.

- duration must be between 0 and 10. For hold\_\allowbreak{}key and wait it is simulated world time under the token clock; hold\_\allowbreak{}key defaults to 1 and wait without duration waits for the next observation. For move,\allowbreak{} mouse\_\allowbreak{}down/up,\allowbreak{} and drag it controls executor-side pointer interpolation only; GUI execution time is excluded from simulated time.

- hold\_\allowbreak{}key keeps every key in the '+'-joined chord down for its full declared duration. A later action does not cancel the hold; another hold on the same key can only extend it. Keys are released automatically at expiry and all held input is released when the episode ends. A held key auto-repeats like a physical keyboard (first repeat after 0.5 s,\allowbreak{} then every 0.05 s),\allowbreak{} so a game that moves one step per key press keeps stepping for as long as the key is held.

- key.repeat is optional,\allowbreak{} defaults to 1,\allowbreak{} and must be an integer from 1 to 100.

- Click variants,\allowbreak{} scroll,\allowbreak{} and drag may include "modifier": a '+'-joined combination of "shift",\allowbreak{} "ctrl",\allowbreak{} "alt",\allowbreak{} and "super". Modifiers apply only for that action.

- mouse\_\allowbreak{}down remains active until the matching mouse\_\allowbreak{}up; use drag for an ordinary complete drag. For a gesture that must react to changing frames,\allowbreak{} use mouse\_\allowbreak{}down,\allowbreak{} then move actions on later observations,\allowbreak{} then mouse\_\allowbreak{}up; each step may use that observation's current marks.

ACTION OUTPUT

Use your configured or native reasoning internally. Your entire visible response must be exactly one valid action from the active action space. Output the action and its arguments only: no reasoning,\allowbreak{} commentary,\allowbreak{} labels,\allowbreak{} wrappers,\allowbreak{} tags,\allowbreak{} Markdown,\allowbreak{} ellipses,\allowbreak{} or placeholders. The action is final and immediately executable. All generated output tokens count. The adapter accepts common formatting wrappers when it can recover exactly one unambiguous valid action. If it finds no valid action or multiple different actions,\allowbreak{} the harness executes WAIT.

\textless{}CLOCK\_\allowbreak{}INSTRUCTION\textgreater{}

TASK: \textless{}TASK\_\allowbreak{}INSTRUCTION\textgreater{}
\end{interactionprompt}
\vspace{-6pt}
\begin{interactionprompt}[InteractionBlue]{Prompt~\ref{prompt:game-role}: Game domain instruction}
\refstepcounter{rtbenchprompt}\label{prompt:game-role}
You are playing a live browser game. Use only the visible game controls and keep acting toward the TASK goal. Do not open a terminal,\allowbreak{} editor,\allowbreak{} source code,\allowbreak{} developer tools,\allowbreak{} or test harness,\allowbreak{} do not reload the page,\allowbreak{} and do not modify the game. Prefer hold\_\allowbreak{}key with a short duration for continuous movement instead of many discrete taps. Keep playing until the environment ends the episode.
\end{interactionprompt}
\vspace{-6pt}
\begin{interactionprompt}[InteractionBlue]{Prompt~\ref{prompt:web-role}: Web domain instruction}
\refstepcounter{rtbenchprompt}\label{prompt:web-role}
You are operating a live web task. The page and server state may change while you reason,\allowbreak{} so use the latest visible state and complete the TASK through the visible UI. Do not inspect or edit source code,\allowbreak{} call hidden APIs,\allowbreak{} or use developer tools or terminal shortcuts. The environment determines when the episode ends; use wait when no further UI interaction is needed.
\end{interactionprompt}
\vspace{-6pt}
\begin{interactionprompt}[InteractionAmber]{Prompt~\ref{prompt:token-clock}: Token-clock instruction}
\refstepcounter{rtbenchprompt}\label{prompt:token-clock}
TOKEN CLOCK: observations are scheduled at f=1 frames per simulated second. Physical inference and GUI execution latency do not advance simulated time. Produce at most one action per observation. B=64 charged tokens/frame; kappa=B*f=64 charged tokens per simulated second. After n provider/native output tokens,\allowbreak{} an action lands n/kappa simulated seconds after the observation. B is a conversion rate,\allowbreak{} not a maximum response length. A decision may use more than B tokens,\allowbreak{} cross scheduled frame boundaries,\allowbreak{} and miss the corresponding observations. Finishing early does not provide an observation before the next scheduled frame boundary. The task ends at its fixed simulated-time limit T,\allowbreak{} with at most kappa*T charged tokens over the episode. Executed actions and their outcomes are retained in your context.
\end{interactionprompt}
\vspace{-6pt}
\begin{interactionprompt}[InteractionAmber]{Prompt~\ref{prompt:paused-clock}: Paused-environment instruction}
\refstepcounter{rtbenchprompt}\label{prompt:paused-clock}
TOKEN CLOCK: observations are scheduled at f=1 frames per simulated second. Physical inference and GUI execution latency do not advance simulated time. Produce at most one action per observation. B is unlimited: output length does not affect simulated time,\allowbreak{} so each action lands at the tick it observed. The environment then advances exactly one frame interval (1/f seconds),\allowbreak{} and the next observation includes the action's effect.
\end{interactionprompt}
\vspace{-6pt}
\begin{interactionprompt}[InteractionAmber]{Prompt~\ref{prompt:wall-clock}: Wall-clock instruction}
\refstepcounter{rtbenchprompt}\label{prompt:wall-clock}
WALL CLOCK: observations are scheduled at 1 fps. Environment time follows physical elapsed time. Produce at most one action per observation and use execution feedback to track which actions were applied.
\end{interactionprompt}

\clearpage
\endgroup

\end{document}